\documentclass[10pt,twocolumn,letterpaper]{article}

\usepackage[pagenumbers]{cvpr} 

\definecolor{cvprblue}{rgb}{0.21,0.49,0.74}
\usepackage[pagebackref,breaklinks,colorlinks,allcolors=cvprblue]{hyperref}
\usepackage{url}            
\usepackage{booktabs}       
\usepackage{amsfonts}       
\usepackage{nicefrac}       
\usepackage{microtype}      
\usepackage{xcolor}         
\usepackage{graphicx}
\usepackage{amsmath}
\usepackage{multirow}
\usepackage{bbm}
\usepackage{pifont}
\usepackage{amsmath}
\usepackage{wrapfig}
\newcommand{\myparagraph}[1]{\smallskip\noindent\textbf{#1}}
\newcommand{\cmark}{\ding{51}}%
\newcommand{\xmark}{\ding{55}}%

\def\paperID{11703} 
\def\confName{CVPR}
\def\confYear{2026}

\title{Learning to Upscale 3D Segmentations in Neuroimaging}

\author{
  Xiaoling Hu$^{1,\dagger}$, Peirong Liu$^{1,2}$, Dina Zemlyanker$^{1}$, Jonathan Williams Ramirez$^{1}$, \\ Oula Puonti$^{1,3}$,
  Juan Eugenio Iglesias$^{1,4,5}$ \\
  \\
  $^{1}$Massachusetts General Hospital and Harvard Medical School \\ 
  $^{2}$Department of ECE, Johns Hopkins University \\ 
  $^{3}$Danish Research Centre for Magnetic Resonance, Copenhagen University Hospital \\
  $^{4}$Hawkes Institute, University College London \\
  $^{5}$Computer Science and AI Laboratory, Massachusetts Institute of Technology \\
}

\begin{document}
\maketitle
\renewcommand{\thefootnote}{\fnsymbol{footnote}}
\footnotetext{$\dagger$ Email: Xiaoling Hu (xihu3@mgh.harvard.edu)}
\begin{abstract}
Obtaining high-resolution (HR) segmentations from coarse annotations is a pervasive challenge in computer vision. Applications include inferring pixel-level segmentations from token-level labels in vision transformers, upsampling coarse masks to full resolution, and transferring annotations from legacy low-resolution (LR) datasets to modern HR imagery. These challenges are especially acute in 3D neuroimaging, where manual labeling is costly and resolutions continually increase. We propose a scalable framework that generalizes across resolutions and domains by regressing signed distance maps, enabling smooth, boundary-aware supervision. Crucially, our model predicts \textbf{one class at a time}, which substantially reduces memory usage during training and inference (critical for large 3D volumes) and naturally supports generalization to unseen classes. Generalization is further improved through training on synthetic, domain-randomized data. We validate our approach on ultra-high-resolution (UHR) human brain MRI ($\sim$100~µm), where most existing methods operate at 1~mm resolution. Our framework effectively upsamples such standard-resolution segmentations to UHR detail. Results on synthetic and real data demonstrate superior scalability and generalization compared to conventional segmentation methods.
Code is available at: \url{https://github.com/HuXiaoling/Learn2Upscale}.

\end{abstract}

\section{Introduction}

A persistent challenge in computer vision lies in bridging the gap between low-cost, coarse-grained annotations and the high-resolution (HR) data produced by modern sensors. This resolution mismatch often manifests when attempting to adapt legacy datasets, often annotated at low resolutions, for use with new HR imagery (a form of domain adaptation~\cite{ghafoorian2017transfer}). It also arises when trying to minimize the intense manual labor required for dense, pixel-perfect labeling, which has spurred research into weakly-supervised methods, from interactive segmentation~\cite{feng2021interactive} to modern prompt-based models~\cite{kirillov2023segment}. Simply upsampling coarse labels or training models on mismatched resolutions typically yields poor results, with blocky, unrealistic boundaries that fail to capture the fine geometric details present in the HR data.

Nowhere is this challenge more extreme than in 3D neuroimaging — for example, human brain MRI, where segmentation is a fundamental task for various downstream applications, including tumor diagnosis and monitoring~\cite{vaidyanathan1997monitoring, mazzara2004brain, logeswari2010improved, kamnitsas2017efficient} and volumetric shape analyses~\cite{hynd1991corpus, galinsky2014automated, goldszal1998image}. For about a decade, deep learning methods like the U-Net~\cite{ronneberger2015u, milletari2016v} have excelled at segmenting standard-resolution scans (1~mm isotropic), for which many labeled atlases and datasets exist~\cite{henschel2020fastsurfer, iglesias2015multi}. However, emerging ultra-high-resolution (UHR) imaging (e.g., \emph{ex vivo} MRI, Hip-CT) now captures data at a much higher resolution. For example, 100-micron \emph{ex vivo} MRI is becoming a commodity~\cite{khandelwal2024automated}, but the 1000-fold increase in volumetric data renders existing segmentation pipelines~\cite{akkus2017deep, grovik2020deep, milletari2016v, chen2021transunet, butoi2023universeg} obsolete for two key reasons. First, computationally, a standard 3D U-Net cannot be easily applied to these massive volumes without exceeding GPU memory limits. Second, data-wise, manually creating new, dense 3D annotations at this 100~µm scale is prohibitively expensive and labor-intensive, making fully-supervised approaches impractical. The field is thus left with a critical need: a method that can leverage the vast repository of existing 1~mm segmentations to produce detailed, accurate results on new 100~µm scans, all while remaining computationally tractable.

In this paper, we propose a scalable and generalizable framework designed to address the resolution gap and computational burden of this task, while accounting for domain shift. To bridge the resolution gap and produce high-quality boundaries, our method moves away from predicting discrete segmentation masks. Instead, it learns to regress per-class signed distance maps (SDFs). This continuous representation is ideal for our upsampling task, as it enables smooth, boundary-aware supervision and encourages the network to infer a geometrically plausible surface, even from coarse, low-resolution (LR) guidance. By optimizing the network to predict a continuous field, we avoid the ``blocky’’ artifacts of discrete upsampling. We further regularize this process with a \emph{gradient norm} loss to enforce sharp boundary properties and a \textit{total variation} (TV) loss to promote local smoothness.

Crucially, to tackle the computational intractability of UHR 3D volumes, we introduce a novel \emph{scalable class-conditional segmentation} (SCCS) mechanism. Rather than attempting to predict all anatomical structures at once in a (potentially huge) multi-channel output, our model is conditioned to predict one class at a time. This simple but powerful design dramatically reduces the memory footprint during training and inference, as the model only needs to hold a single-channel output map in memory. This strategy is the key to making end-to-end training on full UHR volumes feasible. As an added benefit, this one-at-a-time approach naturally supports generalization to unseen anatomical classes, as the model learns a general, class-agnostic segmentation function that is simply guided by the class-specific condition.

We validate our approach on both synthetic data and a challenging real-world dataset of UHR human brain MRI. Our framework effectively upsamples standard 1~mm segmentations to UHR detail, demonstrating superior scalability, accuracy, and generalization compared to conventional segmentation methods. Our key contributions are:
\begin{enumerate}
\item A general, geometry-aware framework for upsampling coarse 3D segmentations to UHR by regressing regularized signed distance maps (SDFs).
\item A \textit{scalable class-conditional segmentation} (SCCS) mechanism that predicts one class at a time, drastically reducing memory consumption and enabling wide generalization to unseen classes, \emph{without retraining or finetuning}.
\item We are the first, to our knowledge, to achieve successful upsampling of UHR brain MRI segmentations from 1~mm to $\sim$100~µm resolution using a deep learning model.
\end{enumerate}

\section{Related Work}

\myparagraph{Deep Learning for Medical Image Segmentation}.
Deep convolutional neural networks (CNNs) have become the state-of-the-art for many segmentation tasks in both natural images~\cite{long2015fully, chen2014semantic, chen2018deeplab, chen2017rethinking, noh2015learning} and the medical domain~\cite{ronneberger2015u, kamnitsas2017efficient}. In medicine, the U-Net architecture~\cite{ronneberger2015u} and its 3D variants~\cite{milletari2016v, isensee2021nnu, chen2021transunet, butoi2023universeg} are dominant. For brain MRI, specifically, many approaches have been proposed, from multi-atlas methods~\cite{iglesias2015multi} to patch-based~\cite{havaei2017brain} and whole-volume architectures like QuickNAT~\cite{roy2019quicknat} and FastSurfer~\cite{henschel2020fastsurfer}. More recently, Vision Transformers (ViTs)~\cite{dosovitskiy2020image} and their specialized variants have gained traction, using self-attention to capture long-range contextual dependencies, leading to models like TransUNet~\cite{chen2021transunet} and nnFormer~\cite{zhou2021nnformer}. While successful, these models are often extremely memory hungry, as the quadratic complexity of self-attention on large feature maps exacerbates the computational burden for massive 3D volumes. Furthermore, their strength lies in modeling long-range global context; this is a feature that is less critical in our specific task, where the focus is on fine-grained boundary detail for an individual, conditioned class, leveraging a coarse spatial prior. Our goal is not to improve the initial coarse segmentation's global context, but rather to upscale its boundaries to UHR detail in a memory-efficient manner. Thus, our approach prioritizes a scalable architecture capable of local geometric refinement, and is thus more suited to CNNs.

Despite this success, these methods face significant challenges when applied to UHR data. First, the massive voxel count of UHR volumes  (often $\sim10^{10}$ voxels)  imposes prohibitive memory and computational burdens for standard whole-volume models. Second, these models are typically trained on dense, full-resolution labels, which are non-existent for UHR \emph{ex vivo} brain scans. These limitations necessitate a new paradigm that is both computationally efficient and capable of learning from coarse or LR supervision.

\myparagraph{Scalable and Conditional Segmentation}.
As dataset resolutions and class numbers increase, scalable segmentation has become a key research area. Traditional methods that predict all classes simultaneously in a multi-channel output mask scale poorly. To address this, recent work has explored conditional or class-wise strategies. In panoptic segmentation, models separate class-agnostic instance prediction from class-level semantics~\cite{kirillov2019panoptic}. More recently, prompt-based models like the Segment Anything Model (SAM)~\cite{kirillov2023segment} have shown remarkable generalization by conditioning on user-provided points, boxes, or masks. In the medical field, modular networks~\cite{valindria2018multi, zhou2019models} and class-conditional approaches~\cite{sun2022inferring, chen2024cpm, zhang2024data} have been proposed to segment one structure at a time, which can improve performance on rare classes and allow for generalization. This philosophy is also shared by incremental~\cite{cermelli2020modeling, garg2022multi, ganea2021incremental} and few-shot/interactive~\cite{tang2021recurrent, ouyang2022self, feng2021interactive} segmentation, which leverage conditioning to enable label-efficient learning. Our work builds directly on this idea, using a class-conditional framework as the key to unlocking computational scalability for massive 3D volumes.

\myparagraph{Domain Randomization}.
A critical challenge in neuro image analysis (especially uncalibrated modalities like MRI) is the lack of generalization across diverse scanning platforms, acquisition protocols, etc. Recent work in image segmentation~\cite{billot2023synthseg,billot2023robust, liu2024brain, hu2025learn2synth}, registration~\cite{hoffmann2021synthmorph, gopinath2024registration, hu2024hierarchical}, and super-resolution~\cite{iglesias2023synthsr} has shown that \textit{domain randomization}~\cite{tobin2017domain} offers a powerful solution. This approach involves training networks exclusively on synthetic data generated from simple anatomical atlases. Crucially, at every training iteration, imaging parameters such as contrast, noise, spatial resolution, and field inhomogeneity are aggressively randomized. This simple strategy forces the network to learn features that are invariant to these common domain shifts. The result is a highly robust network capable of segmenting unseen, real-world images ``out of the box,'' with no need for fine-tuning or adaptation on the target domain~\cite{gopinath2024synthetic}. This paradigm shift towards training on randomized synthetic data is a major inspiration for our work, as it underpins our model's ability to generalize across the massive resolution gap between LR coarse labels and HR target imagery.

\myparagraph{Geometry-Aware Representations and SDFs}.
Most segmentation networks are supervised with binary masks and optimized with cross-entropy or Dice loss. An alternative is to use geometry-aware representations like distance transform maps (DTMs)~\cite{saha2002fuzzy, ma2020distance}. Signed Distance Functions (SDFs)~\cite{osher2003constructing}, which distinguish the interior and exterior of an object, have been widely used in 3D shape representation~\cite{park2019deepsdf, chibane2020neural, chou2022gensdf} and implicit neural rendering~\cite{jiang2020sdfdiff}. Regressing SDFs instead of masks has also been shown to improve boundary delineation in segmentation tasks~\cite{wang2020deep, audebert2019distance, bogensperger2025flowsdf, chen2023unsupervised}. By regressing a continuous SDF, the network can be supervised to learn implicit object boundaries, making it a natural choice for our task of inferring HR details from coarse, LR guidance.

\section{Methods}

\myparagraph{Preliminaries}.
We consider supervised segmentation of UHR brain images, learned from triplets \((I_H, S_{l}, S_{H})\).
Here, \(I_H \in \mathbb{R}^{1 \times D \times H \times W}\) is the UHR input, \(S_{H} \in \mathbb{R}^{C \times D \times H \times W}\) is the corresponding UHR ground-truth (one-hot encoded, $S_H \in \{0,1\}$) segmentation with $C$ classes, and \(S_{l} \in \mathbb{R}^{C \times D' \times H' \times W'}\) (with $D'<D, H'<H, W'<W$) is a coarse, LR segmentation that provides spatial guidance.
Our goal is to learn a network \(\mathbf{F}_{\boldsymbol{\theta}}\) with parameters \(\boldsymbol{\theta}\) that produces an HR prediction conditioned on the LR reference:
\[
\hat{S}_H = \mathbf{F}_{\boldsymbol{\theta}}(I_H \mid S_l), \hat{S}_H \in [0,1]. 
\]
When the conditioning is clear from context, we omit it and write \(\hat{S}_H = \mathbf{F}_{\boldsymbol{\theta}}(I_H)\) for brevity.

The remainder of this section is organized as follows. In~\Cref{sec:naive_lowres_guidance}, we introduce the baseline setting of supervised segmentation using LR annotations as auxiliary spatial guidance, which serves as the basis of our framework.~\Cref{sec:distance} describes our geometry-aware formulation based on regressing signed distance transform maps instead of discrete segmentation labels, enabling smooth and boundary-sensitive supervision. In~\Cref{sec:scalable_segmentation}, we present the proposed SCCS strategy, which allows the model to efficiently scale to a large number of anatomical structures and generalize to unseen classes through per-class conditional training.

\subsection{Supervised Segmentation with LR Guidance}
\label{sec:naive_lowres_guidance}

\subsubsection{Baseline: Direct Supervised Segmentation}
\label{sec:naive}
A straightforward solution for segmenting UHR brain MRI volumes is to directly train a segmentation network that maps the full-resolution image \( I_H \in \mathbb{R}^{1 \times D \times H \times W} \) to its corresponding voxel-wise label map \( S_H \in \mathbb{R}^{C \times D \times H \times W} \). The network is typically optimized using a standard voxel-level loss, such as multi-class cross-entropy or Dice loss, to predict the full-resolution segmentation.  

While simple in principle, this \textit{fully supervised paradigm} suffers from severe computational and practical constraints. The vast spatial resolution of \( I_H \) entails extremely high GPU memory demands during both training and inference, making end-to-end optimization nearly infeasible on commodity hardware. Consequently, most existing approaches rely on patch-based sampling or sliding-window strategies, which reduce the field of view and compromise anatomical context~\cite{zeng2024segmentation,khandelwal2024automated}. This loss of contextual information often leads to fragmented predictions and poor global consistency across patches.  

Moreover, acquiring voxel-level annotations at full resolution (\( S_H \)) is highly expensive, requiring extensive manual labor and anatomical expertise. In practice, such densely labeled datasets exist only for small datasets or a small number of labels~\cite{khandelwal2024automated}, hindering generalization and scalability. This motivates the question of our work: \textit{Can we achieve fine-grained, geometry-aware segmentation of UHR brain MRIs with additional coarse or LR supervision?}  

We begin with this naive, fully supervised setup as a baseline, then progressively enhance it with new mechanisms that \textit{(i)}~exploit weak, LR labels, \textit{(ii)}~embed geometric structure into learning, and \textit{(iii)}~scale to many classes efficiently. Also, \textit{domain randomization} strategy is employed during the training to achieve generalizability.

\subsubsection{LR Segmentation as Auxiliary Guidance}
\label{sec:low_guide}

\begin{figure*}[t]
\begin{center}
\includegraphics[width=0.9\textwidth]{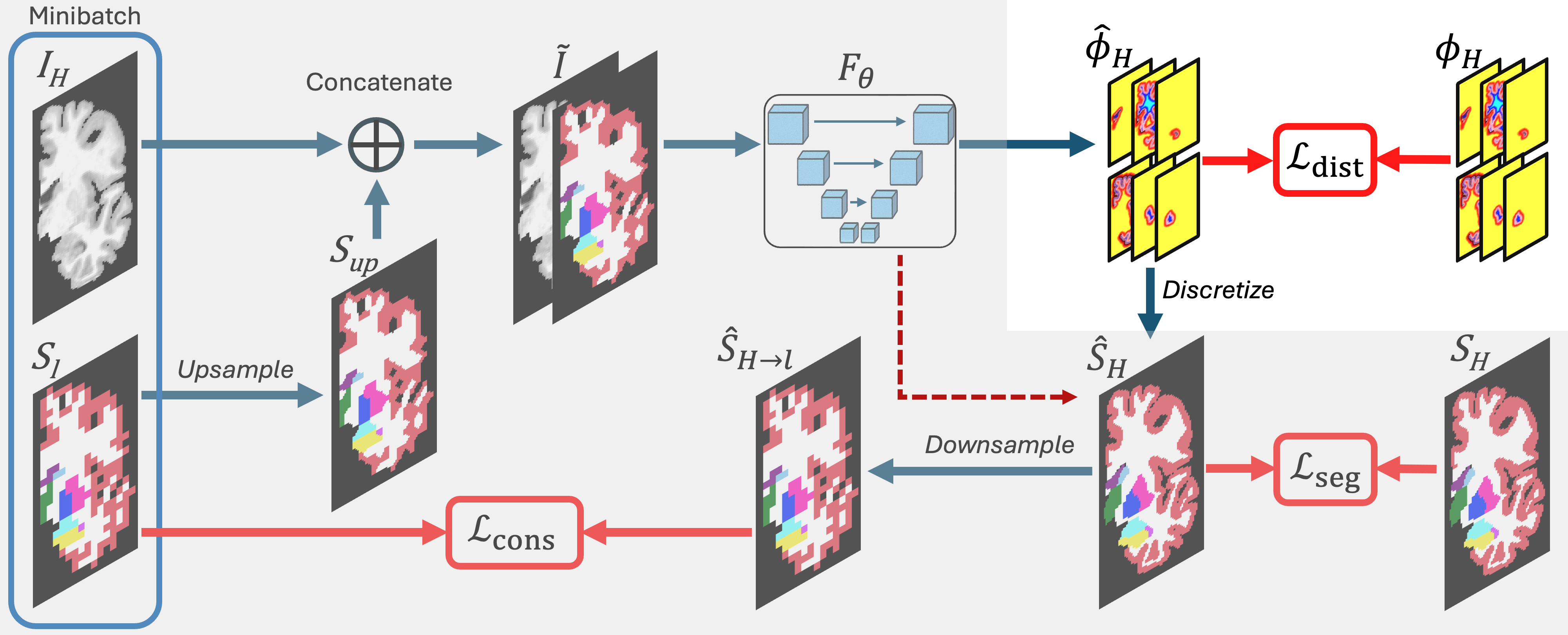}
\end{center}
\caption{Overview of the proposed LR-guided and distance-based representation framework. In addition to the standard segmentation loss $\mathcal{L}_{\text{seg}}$, we introduce a cross-resolution consistency term $\mathcal{L}_{\text{cons}}$ (see \Cref{sec:naive} and \Cref{eq:naive}), illustrated in the shaded region. We further regress signed distance maps ($\hat{\phi}_H$) to enable a geometry-aware representation (\Cref{sec:distance} and \Cref{loss:sdf_loss}), as depicted in the full workflow.}
\label{fig:fw_dist}
\end{figure*}

In contrast to the conventional HR-only paradigm, we propose to \textit{leverage automatically obtained LR segmentations} as auxiliary supervisory signals. In human brain MRI, LR anatomical maps \( S_l \) can be efficiently generated using robust and general-purpose segmentation tools such as SynthSeg~\cite{billot2023synthseg}, which require no manual annotation. Although coarse, these maps capture meaningful global spatial priors that can be propagated into UHR training.  

Our approach integrates \( S_l \) into training in two complementary ways, forming the first key contribution of our framework, \textit{LR-guided supervision}.  

\myparagraph{Prior-Guided Input Augmentation}. 
We first upsample the LR segmentation to the HR space via trilinear interpolation on the one-hot encoding (\Cref{fig:fw_dist}, left):
\[
S_{\text{up}} = \text{Upsample}(S_l),
\]
and concatenate it with the raw MRI volume:
\[
\tilde{I} = \text{Concat}(I_H, S_{\text{up}}) \in \mathbb{R}^{(1 + C) \times D \times H \times W}.
\]
This augmented input explicitly encodes semantic context from \( S_l \), enabling the model to localize fine structures while retaining a global understanding of brain anatomy. Intuitively, \( S_{\text{up}} \) provides a coarse anatomical atlas that conditions the network toward more plausible segmentation hypotheses. While upscaling up front is less memory efficient than in later stages, it enables compatibility with LR inputs of any size.

\myparagraph{Cross-Resolution Semantic Consistency}.
In parallel, we enforce a cross-resolution alignment between predicted HR segmentations and their LR counterparts. The model output \(\hat{S}_H = \mathbf{F}_{\boldsymbol{\theta}}(\tilde{I})\) is downsampled to the coarse scale:
\[
\hat{S}_{H \to l} = \text{Downsample}(\hat{S}_H),
\]
and compared against the reference \( S_l \) using a Dice consistency loss:
\[
\mathcal{L}_{\text{cons}} = \text{Dice}(\hat{S}_{H \to l}, S_l).
\]
The total loss is thus:
\begin{equation}
\label{eq:naive}
\mathcal{L}_{\text{total}} = \mathcal{L}_{\text{seg}} + \lambda_{\text{cons}}\mathcal{L}_{\text{cons}},
\end{equation}
where \( \lambda_{\text{cons}} \) controls the strength of cross-resolution regularization. This auxiliary loss effectively aligns the model’s HR predictions with coarse global priors, ensuring semantic coherence across scales.

\paragraph{Challenges and Motivation for Continuous Representations.}
Patch-based training introduces inevitable alignment issues between UHR image regions and corresponding LR labels, particularly when spatial transformations are used in augmentation. Small misalignments can corrupt consistency supervision. More fundamentally, direct voxel-level classification imposes rigid, discrete boundaries, making optimization unstable and insensitive to geometric smoothness. Such formulations often yield noisy, discontinuous, or topologically inconsistent segmentations, an undesirable property when reconstructing fine anatomical interfaces.  

To overcome these limitations, we reformulate the segmentation task from a \textit{categorical labeling problem} into a \textit{continuous geometric regression problem}, leading to our \textit{geometry-aware signed distance transform learning}.

\subsection{Learning Geometry-Aware Representations via Signed Distance Transforms}
\label{sec:distance}


\myparagraph{Definition}. Given a 3D multi-class segmentation map $S : \Omega \subset \mathbb{R}^3 \rightarrow \mathbb{R}^C$, where each voxel $x \in \Omega$ is assigned a class label, the signed distance map $\phi^c : \Omega \rightarrow \mathbb{R}$ for each class $c \in \{1, \dots, C-1\}$ is defined as:
\[
\phi^c(x) =
\begin{cases}
  -\min\limits_{y \in \partial \Omega^c} \|x - y\|_2, & \text{if } x \in \Omega^c \\
  \phantom{-}\min\limits_{y \in \partial \Omega^c} \|x - y\|_2, & \text{otherwise,}
\end{cases}
\]
where, $\Omega^c = \{x \mid S(x,y,z,c) = 1\}$ is the foreground region for class $c$, $\partial \Omega^c$ denotes the boundary of $\Omega^c$, and $\|\cdot\|_2$ is the Euclidean distance in 3$D$ space. The full multi-class signed distance map can be represented as a tensor $\phi \in \mathbb{R}^{C \times D \times H \times W}$, where $\phi^c$ corresponds to the distance map for class $c$.

Rather than predicting discrete voxel labels, we train the network to regress these continuous signed distance maps. This design introduces several distinct advantages over classical segmentation. First, SDFs represent spatial proximity to anatomical boundaries, capturing both interior and exterior geometry. This continuous representation yields smoother gradients and inherently encodes shape priors. Second, Distance regression provides stable optimization even in regions of partial volume or fuzzy boundaries, an essential property for submillimeter brain structures. Third, because distance fields vary smoothly across space, they naturally tolerate small misalignments or label noise, particularly when supervision comes from LR 
This constitutes a significant shift from discrete classification toward geometry-aware continuous learning for UHR segmentation.

\paragraph{Learning Formulation.}
The network $\mathbf{F}_{\boldsymbol{\theta}}$ predicts UHR distance maps:
\[
\hat{\phi}_H = \mathbf{F}_{\boldsymbol{\theta}}(\tilde{I}) \in \mathbb{R}^{C \times D \times H \times W},
\]
and is supervised with an $\ell_1$ regression loss:
\[
\mathcal{L}_{\text{dist}} = \|\hat{\phi}_H - \phi_H\|_1.
\]
We convert $\hat{\phi}_H$ to probabilistic segmentations via a temperature-controlled softmax over negative distances:
\[
\hat{S}_H^c(v) = \frac{\exp(-\hat{\phi}_H^{c}(v) / \tau)}{\sum_{c'} \exp(-\hat{\phi}_H^{c'}(v) / \tau)}.
\]
Here, $\phi_H^c(v)$ denotes the predicted signed distance at voxel $v$ for class $c$, $\tau$ is the temperature parameter (controls the sharpness of the distance-to-probability mapping), $\hat{S}_H^c(v) \in [0,1]$ is the probability that voxel $v$ belongs to class $c$, and the output $\hat{S}_H \in \mathbb{R}^{C \times D \times H \times W}$ is a probabilistic segmentation map. This mapping smoothly bridges continuous distances and categorical probabilities, ensuring differentiability and interpretability.

\paragraph{Geometry-Aware Regularization.}
To further enhance the geometric fidelity of learned SDFs, we introduce two additional regularizers:
\[
\mathcal{L}_{\nabla} = \frac{1}{|\Omega|} \sum_{v} (\|\nabla \hat{\phi}_H(v)\|_2 - 1)^2,
\]
\[
\mathcal{L}_{\text{TV}} = \frac{1}{|\Omega|} \sum_{v} \sqrt{\sum_{i=1}^{3} (\nabla_i \hat{\phi}_H(v))^2}.
\]
The gradient norm term ($\mathcal{L}_{\nabla}$) enforces the unit-gradient property of true SDFs~\cite{ma2023towards}, while total variation (TV, $\mathcal{L}_{\text{TV}}$) regularization suppresses spurious local noise and promotes smooth boundary transitions. Together, they impose strong geometric priors that stabilize training and improve generalization.

Finally, we incorporate the cross-resolution consistency term from~\Cref{sec:low_guide}:
\[
\mathcal{L}_{\text{cons}} = \text{Dice}(\text{Downsample}(\hat{S}_H), S_l),
\]
and define the complete loss as:
\begin{equation}
\label{loss:sdf_loss}
\mathcal{L}_{\text{total}}^{geo} =
\mathcal{L}_{\text{dist}} +
\lambda_{\text{gn}}\mathcal{L}_{\nabla} +
\lambda_{\text{tv}}\mathcal{L}_{\text{TV}} +
\lambda_{\text{cons}}\mathcal{L}_{\text{cons}}.
\end{equation}
The architecture is illustrated in~\Cref{fig:fw_dist}. At inference, segmentation labels are obtained via $\hat{S}_H(v) = \arg\min_c \hat{\phi}_H^c(v)$.

This formulation bridges discrete semantic segmentation and continuous shape modeling. It encodes boundary geometry directly within the learning target, significantly improving smoothness, robustness, and topological integrity for UHR brain segmentation. In combination with LR consistency, this produces fine, anatomically coherent predictions even with limited annotations.

\subsection{Scalable Class-Conditional Segmentation (SCCS)}
\label{sec:scalable_segmentation}

\begin{figure}[ht]
\begin{center}
\includegraphics[width=0.49\textwidth]{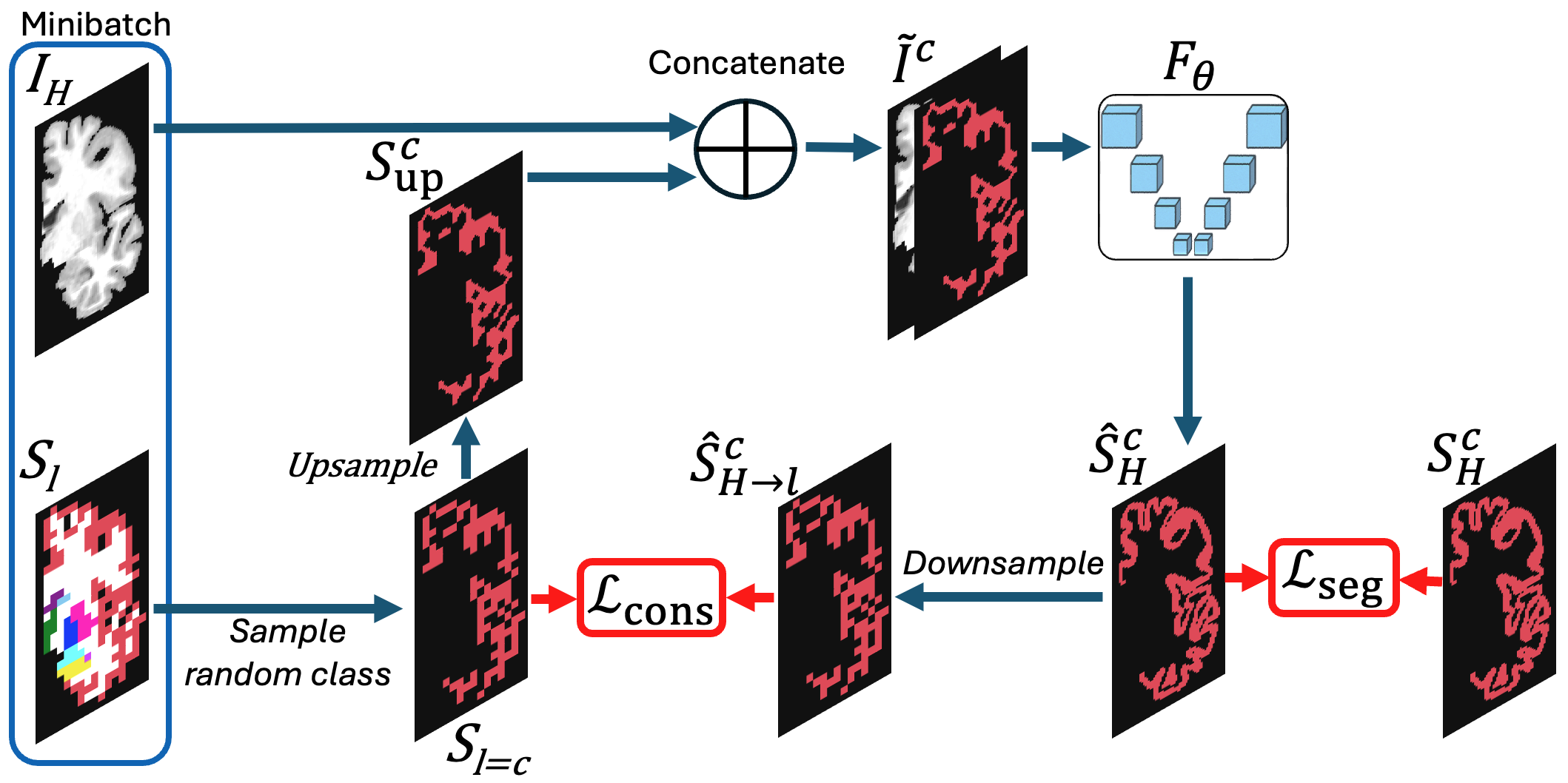}
\end{center}
\caption{Illustration of the SCCS framework. At each training step, the model focuses on a single class, substantially reducing memory footprint and allowing flexible extension to new anatomical structures. For the selected class $c$, the segmentation loss $\mathcal{L}_{\text{seg}}^c$ and the cross-resolution consistency loss $\mathcal{L}_{\text{cons}}^c$ are combined into a total objective $\mathcal{L}_{\text{total}}^c$ (\Cref{eq:condition_loss}), which supervises the training of the entire network.}
\label{fig:fw_scalable}
\end{figure}

While the distance-based representation addresses geometric and boundary limitations, scaling segmentation to a large number of structures introduces additional computational bottlenecks. Standard multi-class networks must allocate one output channel per label, making them prohibitively memory-heavy for UHR volumes containing dozens or hundreds of anatomical regions.  

To overcome this, we propose \textit{scalable class-conditional segmentation} (SCCS), which reformulates multi-class segmentation as a collection of class-specific subproblems.

\paragraph{Class-Conditional Training.}
Instead of predicting all classes jointly, the model learns to segment one class at a time, conditioned on a class-specific input. At each iteration, we randomly sample a target class \( c \) and extract its binary mask from the LR reference \( S_l \):
\[
S^c_{\text{up}} = \text{Upsample}(\mathbbm{1}_{S_l = c}),
\]
then concatenate it with the input image:
\[
\tilde{I}^c = \text{Concat}(I_H, S^c_{\text{up}}) \in \mathbb{R}^{(1+1) \times D \times H \times W}.
\]
This conditioning localizes the model’s attention to anatomically relevant regions, simplifying learning and improving sample efficiency. The model predicts a binary segmentation map $\hat{S}_H^c = \mathbf{F}_{\boldsymbol{\theta}}(\tilde{I}^c)  \in \mathbb{R}^{1 \times D \times H \times W}$ and is trained via:
\[
\mathcal{L}_{\text{seg}}^c = \text{BCE}(\hat{S}_H^c, \mathbbm{1}_{S_H = c}),
\]
\[
\mathcal{L}_{\text{cons}}^c = \text{BCE}(\text{Downsample}(\hat{S}_H^c), \mathbbm{1}_{S_l = c}).
\]
The total per-class loss is
\begin{equation}
\label{eq:condition_loss}
\mathcal{L}_{\text{total}}^c = \mathcal{L}_{\text{seg}}^c + \lambda_{\text{cons}}\mathcal{L}_{\text{cons}}^c,
\end{equation}
and the whole framework is illustrated in~\Cref{fig:fw_scalable}.

\paragraph{Inference and Scalability.}
At inference time, the model is applied for each individual class \( c \) using a class-conditioned input \( \tilde{I}^{c} \), which includes the image and the class-specific conditioning signal. The model produces a single-channel response map \( \hat{S}^{c} \in \mathbb{R}^{1 \times D \times H \times W} \), indicating the probability that each voxel belongs to class \( c \). After looping through all classes, the individual response maps are stacked to form a multi-class prediction volume. The final multi-class segmentation is reconstructed via:
\[
\hat{S}_H(v) = \arg\max_c \hat{S}^c(v).
\]
This design scales linearly with the number of classes and can seamlessly adapt to unseen anatomical labels by conditioning on their corresponding LR masks, without any retraining or architectural modification.

\myparagraph{Generalizability via Domain Randomization}. For our training paradigms, robust generalization is essential, as both the HR input \(I_H\) and the LR guidance \(S_l\) may originate from diverse scanners and acquisition protocols. To prevent the model from overfitting to a narrow appearance distribution, we incorporate a domain randomization strategy during training. Following recent successes in synthetic neuroimage pipelines~\cite{billot2023synthseg, billot2023robust, hoffmann2021synthmorph, iglesias2023synthsr}, we apply aggressive, stochastic perturbations to image contrast, noise, bias fields, and spatial resolution at each iteration.

This randomized augmentation forces the network to rely on stable structural cues rather than dataset- or scanner-specific appearance details. When combined with the class-conditional input \(\tilde{I}^c\), the model learns class-specific geometry that is inherently invariant to domain shifts. In practice, this significantly improves the model’s robustness across heterogeneous datasets and helps bridge the appearance gap between the coarse LR reference and the UHR target. 

\section{Experiments}

We conduct comprehensive evaluations on both synthetic and real-world datasets with human-level annotations to demonstrate the effectiveness of the proposed methods as well as the effectiveness of the parameter selection.

\myparagraph{Datasets}. We use synthetic data for training, where 400 UHR isotropic scans with a resolution of $\frac{1}{3}$\,mm $\times$ $\frac{1}{3}$\,mm $\times$ $\frac{1}{3}$\,mm, and the other 100 for validation. For evaluation, we employ two test sets: (1) a synthetic test set comprising 100 held-out synthetic volumes, and (2) 20 real scans from the \textit{U01} dataset~\cite{pesce20223d}, each with a single annotated 2D slice. The \textit{U01} surface models were originally generated by converting segmentation probability maps, obtained using a cascaded multi-resolution U-Net~\cite{zeng2024segmentation}, into pseudo T1-weighted scans, followed by surface placement using a modified version of the FreeSurfer \texttt{recon-all} pipeline~\cite{dale1999cortical, fischl1999cortical}. More details are provided in the supplementary material.

\myparagraph{Implementation Details}. We use a standard 3D U-Net~\cite{ronneberger2015u, cciccek20163d} as the backbone for our segmentation. The networks are randomly initialized and trained from scratch. We use the Dice loss~\cite{sudre2017generalised} as segmentation loss and $l_1$ loss as the consistency loss (if applicable) to supervise the training of the network. Adam optimizer~\cite{kingma2014adam} is adopted with a learning rate of $1\times 10^{-3}$. We set the loss weights of $\lambda_{\text{gn}}, \lambda_{\text{tv}}$, and $\lambda_{\text{cons}}$ as 0.1, 0.01, and 1 for all our experiments except for the ablation study sections regarding these parameters. Note that the reported memory usage corresponds to an input size of $192 \times 192 \times 192$ with a batch size of 1. More information is provided as supplementary material.

\myparagraph{Baselines}.  CascadePSP~\cite{cheng2020cascadepsp} adopts a cascaded pyramid refinement strategy in which a coarse segmentation is progressively refined across multiple resolution stages. This hierarchical framework is highly effective for natural image segmentation, where dense pixel-level annotations are available and texture cues are informative. 
CRM~\cite{shen2022high} (Continuous Refinement Model) addresses resolution inconsistencies by explicitly fusing multi-scale feature representations to enhance boundary precision. 

\myparagraph{Evaluation Metrics}. We use Dice score~\cite{zou2004statistical} and the Hausdorff distance (HD95) to report all the performances. The Dice score~\cite{zou2004statistical} is a classical segmentation metric, which measures the overlap between predicted and ground truth masks. The HD95~\cite{huttenlocher1993comparing} calculates the 95th percentile of all boundary distances rather than the absolute maximum. We report both means and standard deviations for all the results, and bolded numbers denote significant differences (t-test, $p=0.05$).

\subsection{Supervised Segmentation with LR Guidance}
We begin by evaluating the proposed supervised segmentation framework under various training configurations, including direct segmentation and distance-transform regression. To contextualize our approach, we compare against two state-of-the-art (SOTA) segmentation models originally developed for HR natural image segmentations: CascadePSP~\cite{cheng2020cascadepsp} and CRM~\cite{shen2022high}. These methods represent strong HR baselines that emphasize multi-scale refinement and cross-resolution fusion, respectively, making them ideal points of comparison for assessing our design in the neuroimaging domain.

\begin{table}[t!]
\centering
\tiny
\setlength{\tabcolsep}{3pt}
\caption{Segmentation results with different settings.}
\begin{tabular}{lcccccc}
\toprule
\multirow{2}{*}{Method} & \multirow{2}{*}{Guidance} & \multicolumn{2}{c}{Synthetic dataset} &  & \multicolumn{2}{c}{U01} \\ \cmidrule{3-4} \cmidrule{6-7} 
 & & Dice$\uparrow$  & HD95 (mm)$\downarrow$ & & Dice$\uparrow$  & HD95 (mm)$\downarrow$  \\ \midrule
Naive seg. (\Cref{sec:naive})  & N/A & 0.754 $\pm$ 0.022 & 0.886 $\pm$ 0.153 & & 0.721 $\pm$ 0.046 & 1.471 $\pm$ 0.204 \\
CascadePSP \cite{cheng2020cascadepsp}  & N/A & 0.747 $\pm$ 0.031 & 0.901 $\pm$ 0.161 & & 0.709 $\pm$ 0.051 & 1.515 $\pm$ 0.228\\
CRM \cite{shen2022high}  & N/A & 0.761 $\pm$ 0.019 & 0.871 $\pm$ 0.146 & & 0.719 $\pm$ 0.049 & 1.501 $\pm$ 0.217 \\
\midrule
Seg. $\hat{S}_H$ (\Cref{sec:low_guide})  &  \multirow{2}{*}{$S_l$} & 0.771 $\pm$ 0.023 & 0.803 $\pm$ 0.124 & & 0.743 $\pm$ 0.037 & \textbf{1.015 $\pm$ 0.198} \\ 
SDF $\hat{\phi}_H$ (\Cref{sec:distance}) &  & \textbf{0.782 $\pm$ 0.015} & \textbf{0.768 $\pm$ 0.106} & & \textbf{0.751 $\pm$ 0.026} & \textbf{0.976 $\pm$ 0.151} \\ 
\bottomrule
\end{tabular}
\label{seg_result}
\end{table}
\myparagraph{Results}. \Cref{seg_result} summarizes the quantitative results across synthetic and real datasets, while qualitative examples are illustrated in~\Cref{fig:qualitative}. Incorporating LR guidance \(S_l\) consistently improves segmentation accuracy and boundary precision across all settings. Notably, the proposed SDF regression yields the highest Dice scores and lowest HD95 distances, surpassing both traditional baselines and direct segmentation models. This demonstrates that learning continuous, geometry-aware representations provides a stronger supervisory signal than discrete voxel classification, particularly when training data are limited.

\begin{figure*}[t]
    \centering
     \setlength{\tabcolsep}{0.1pt}
    \begin{tabular}{ccccccc}
    \includegraphics[width=.125\textwidth]{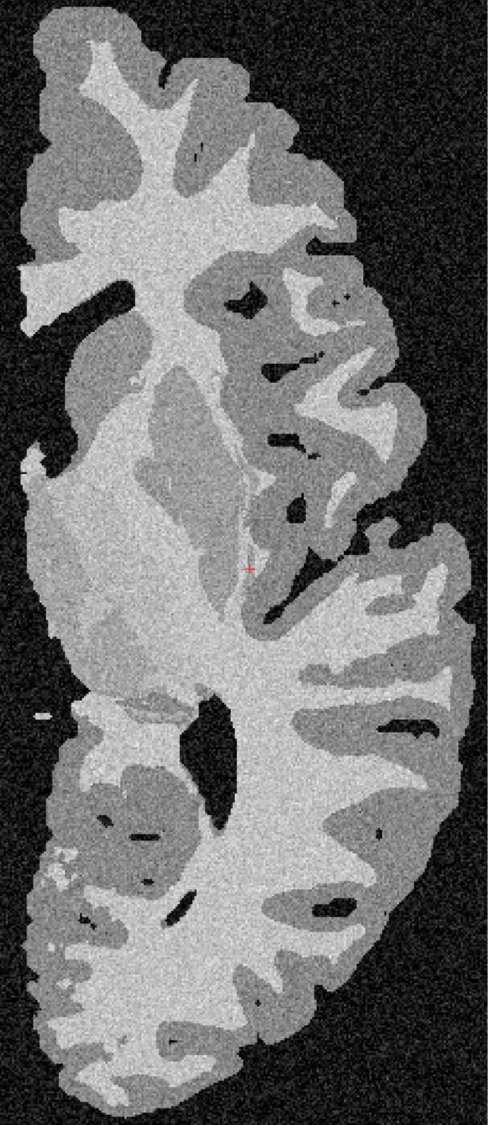} &
    \includegraphics[width=.125\textwidth]{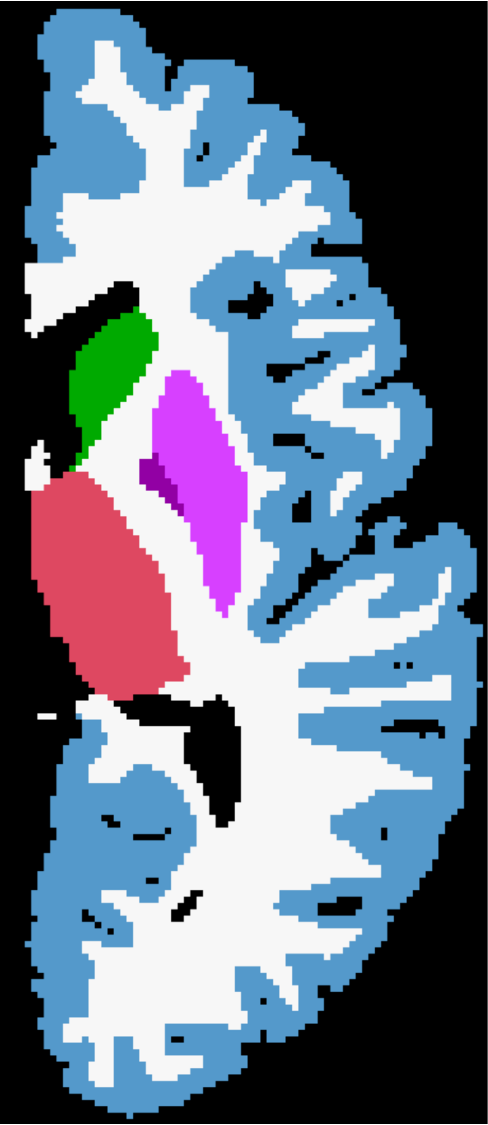} &
    \includegraphics[width=.125\textwidth]{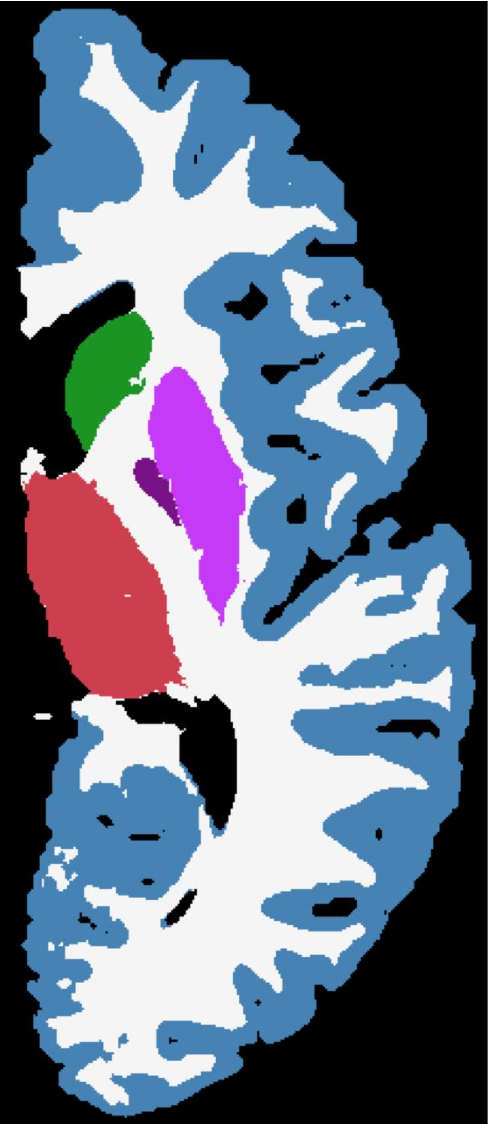} &
    \includegraphics[width=.125\textwidth]{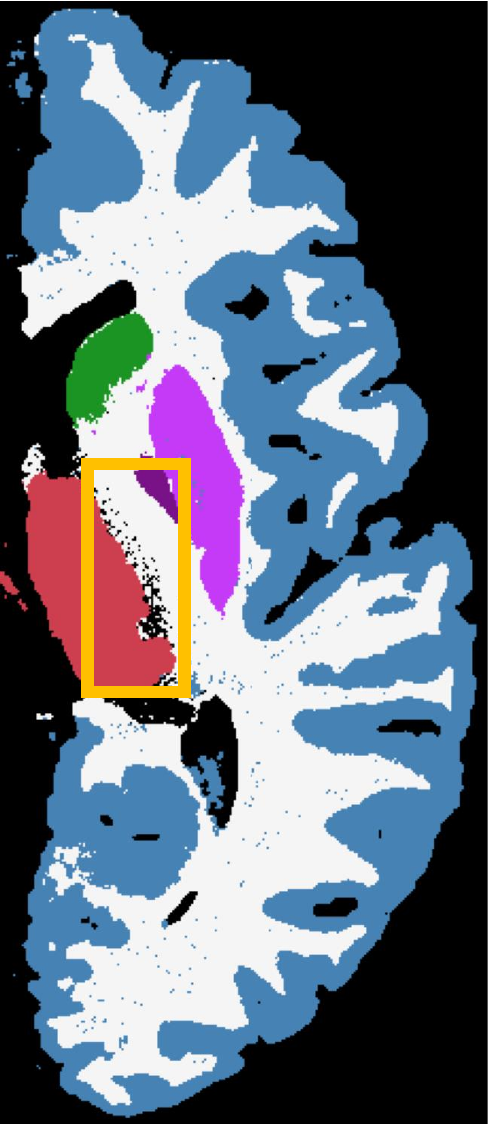} &
    \includegraphics[width=.125\textwidth]{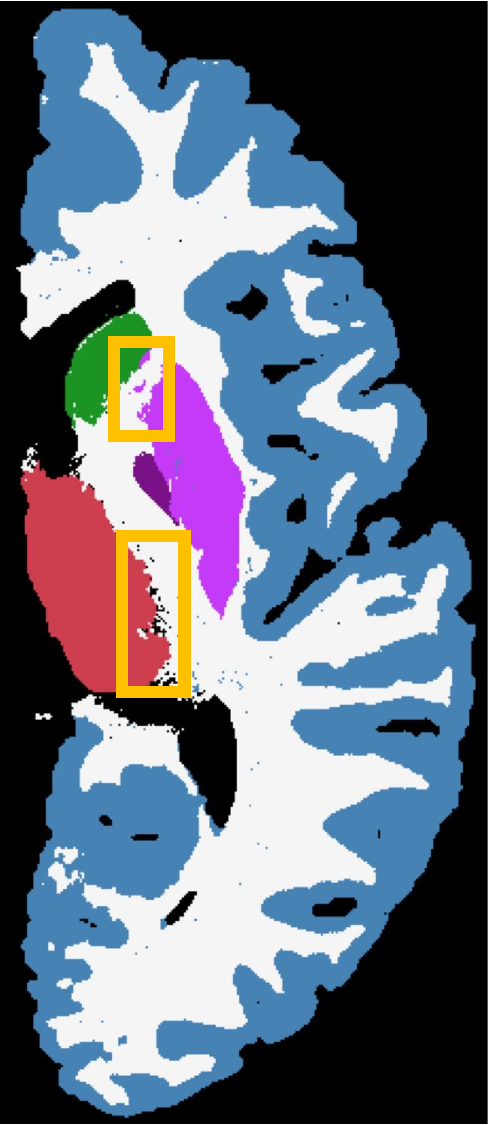} &   
    \includegraphics[width=.125\textwidth]{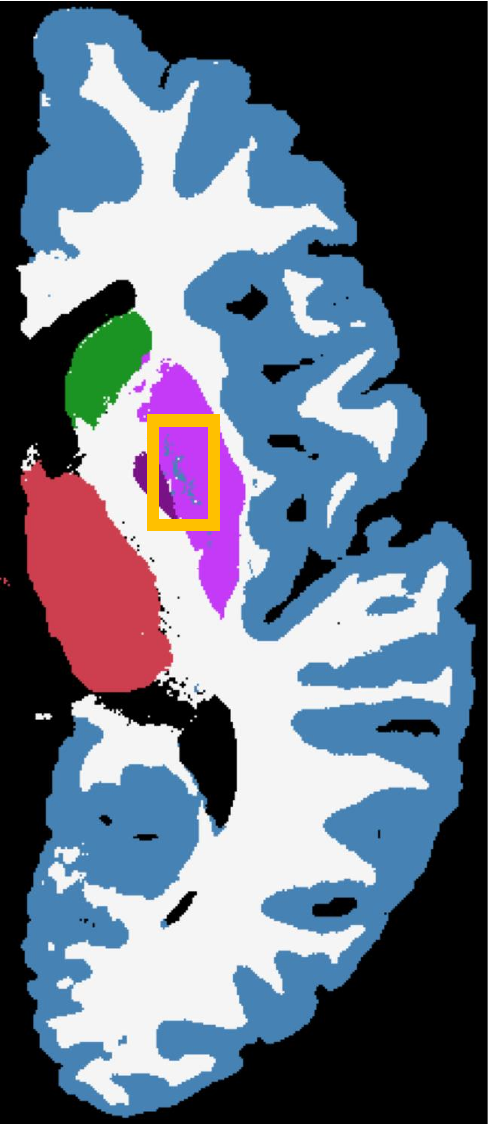} &
    \includegraphics[width=.125\textwidth]{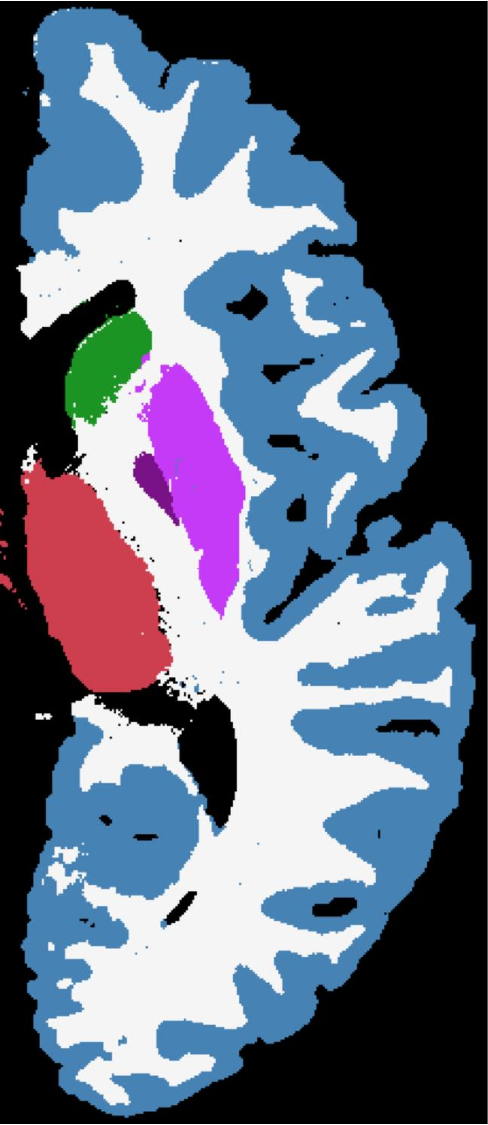} \\
     (a) Input & (b) LR guidance & (c) GT & (d) Naive Seg & (e) Predict $\hat{S}_H$ & (f) Regress $\hat{\phi}_H$ & (g) SCCS  \\
    \end{tabular}
    \caption{Qualitative results. (a-c) show the input, LR guidance, and ground truth (GT). (d-g) show segmentations with different methods. 
    }
    \label{fig:qualitative}
\end{figure*}

Compared with CascadePSP and CRM, our approach shows superior generalization and boundary stability. Whereas the natural-image models focus on iterative refinement of visual textures, our method leverages structural priors and geometric regularization to capture the true anatomical topology of brain regions. This difference is particularly evident in high-curvature or thin cortical regions, where SDF-based learning preserves connectivity and reduces spurious discontinuities. Together, these results highlight that coupling LR anatomical guidance with geometry-aware regression offers a more robust and scalable solution for UHR brain segmentation.

\subsubsection{Ablation Study}
We conduct a series of ablation studies to justify the effectiveness of individual components, as well as the sensitivity to hyperparameters.

\begin{table}[t!]
\centering
\scriptsize
\setlength{\tabcolsep}{4pt}
\caption{Ablation study on $\mathcal{L}_{\text{cons}}$.}
\begin{tabular}{llcc}
\toprule
Method & Setting & Dice$\uparrow$  & HD95 (mm)$\downarrow$ \\ \midrule
\multirow{2}{*}{Predict seg. $\hat{S}_H$} & Seg. as input (w/o $\mathcal{L}_{\text{cons}}$) & 0.765 $\pm$ 0.018 & 0.841 $\pm$ 0.145\\
& Seg. as input (w/ $\mathcal{L}_{\text{cons}}$) & \textbf{0.771 $\pm$ 0.023} & \textbf{0.803 $\pm$ 0.124}\\
\midrule
\multirow{2}{*}{Regress SDF $\hat{\phi}_H$} & Seg. as input (w/o $\mathcal{L}_{\text{cons}}$) & 0.773 $\pm$ 0.016 & 0.826 $\pm$ 0.156\\
& Seg. as input (w $\mathcal{L}_{\text{cons}}$) & \textbf{0.782 $\pm$ 0.015} & \textbf{0.768 $\pm$ 0.106} \\
\bottomrule
\end{tabular}
\label{consistency}
\end{table}

\begin{table*}[ht]
\parbox{.33\linewidth}{
\setlength{\tabcolsep}{10pt}
\centering
\scriptsize
\caption{Ablation study on $\lambda_{\text{gn}}$.}
\begin{tabular}{ccc}
\toprule
$\lambda_{\text{gn}}$  & Dice$\uparrow$  & HD95 (mm)$\downarrow$ \\ \midrule
0 & 0.766 $\pm$ 0.016 & 0.812 $\pm$ 0.124 \\
0.05  & 0.775 $\pm$ 0.018 & 0.826 $\pm$ 0.110 \\
\underline{0.10}  & 0.782 $\pm$ 0.015 & 0.768 $\pm$ 0.106\\
0.15  & 0.778 $\pm$ 0.021 & 0.756 $\pm$ 0.129\\
0.20 & 0.772 $\pm$ 0.017 & 0.801 $\pm$ 0.098 \\
 \bottomrule
\end{tabular}
\label{ablation:loss_gn}
}
\hfill
\parbox{.33\linewidth}{
\setlength{\tabcolsep}{10pt}
\centering
\scriptsize
\caption{Ablation study on $\lambda_{\text{tv}}$.}
\begin{tabular}{ccc}
\toprule
$\lambda_{\text{tv}}$  & Dice$\uparrow$  & HD95 (mm)$\downarrow$ \\ \midrule
0 & 0.768 $\pm$ 0.019 & 0.847 $\pm$ 0.136 \\
0.001  & 0.765 $\pm$ 0.022 & 0.825 $\pm$ 0.101 \\
\underline{0.01}  & 0.782 $\pm$ 0.015 & 0.768 $\pm$ 0.106\\
0.02  & 0.779 $\pm$ 0.010 & 0.771 $\pm$ 0.085 \\
0.05 & 0.766 $\pm$ 0.021 & 0.871 $\pm$ 0.123\\
 \bottomrule
\end{tabular}
\label{ablation:loss_tv}
}
\hfill
\parbox{.33\linewidth}{
\setlength{\tabcolsep}{10pt}
\centering
\scriptsize
\caption{Ablation study on $\lambda_{\text{cons}}$.}
\begin{tabular}{ccc}
\toprule
$\lambda_{\text{cons}}$  & Dice$\uparrow$  & HD95 (mm)$\downarrow$  \\ \midrule
0  & 0.773 $\pm$ 0.016 & 0.826 $\pm$ 0.156\\
0.5  & 0.777 $\pm$ 0.021 & 0.746 $\pm$ 0.090\\
\underline{1.0}  & 0.782 $\pm$ 0.015 & 0.768 $\pm$ 0.106 \\
1.5  & 0.781 $\pm$ 0.019 & 0.790 $\pm$ 0.141\\
2.0 & 0.775 $ \pm$ 0.017 & 0.782 $\pm$ 0.161\\
 \bottomrule
\end{tabular}
\label{ablation:loss_cons}
}
\end{table*}

\myparagraph{Ablation Study on the Consistency Loss Term}. As described earlier, the LR segmentation \(S_l\) not only provides spatial guidance but also enforces semantic alignment between the predicted HR output and its LR reference. To assess the impact of the cross-resolution consistency term \(\mathcal{L}_{\text{cons}}\), we perform an ablation study by removing it from the overall objective. The results in~\Cref{consistency} show a clear and consistent drop in Dice score and an increase in boundary error when \(\mathcal{L}_{\text{cons}}\) is omitted. This confirms that the consistency constraint effectively regularizes the network, promoting spatial coherence and improving generalization across varying resolutions.

\begin{table}[ht]
\setlength{\tabcolsep}{11pt}
\centering
\scriptsize
\caption{Ablation study on loss components.}
\begin{tabular}{ccccc}
\toprule
$\mathcal{L}_{\nabla}$ & $\mathcal{L}_{\text{TV}}$ & $\mathcal{L}_{\text{cons}}$ & Dice$\uparrow$  & HD95 (mm)$\downarrow$ \\ \midrule
 \xmark & \xmark & \xmark & 0.748 $\pm$ 0.025 & 1.023 $\pm$ 0.176\\
 \cmark & \xmark & \xmark & 0.765 $\pm$ 0.018 & 0.937 $\pm$ 0.181\\
 \xmark & \cmark & \xmark & 0.760 $\pm$ 0.012 & 0.895 $\pm$ 0.161\\
 \xmark & \xmark & \cmark & 0.754 $\pm$ 0.021 & 1.123 $\pm$ 0.145\\
 \cmark & \cmark & \xmark & 0.773 $\pm$ 0.016 & 0.826 $\pm$ 0.156 \\
 \cmark & \xmark & \cmark & 0.768 $\pm$ 0.019 & 0.847 $\pm$ 0.136\\
 \xmark & \cmark & \cmark & 0.766 $\pm$ 0.016 & 0.812 $\pm$ 0.124 \\
 \cmark & \cmark & \cmark &  \textbf{0.782 $\pm$ 0.015} & \textbf{0.768 $\pm$ 0.106} \\ \bottomrule
\end{tabular}
\label{ablation:loss_components}
\end{table}

\myparagraph{Ablation Study on Loss Weights}. We next examine the sensitivity of the proposed framework to the weighting coefficients in the total loss formulation (\Cref{loss:sdf_loss}). Specifically, we vary the relative strengths of \(\lambda_{\text{gn}}\), \(\lambda_{\text{tv}}\), and \(\lambda_{\text{cons}}\) as reported in~\Cref{ablation:loss_gn},~\Cref{ablation:loss_tv}, and~\Cref{ablation:loss_cons}. The results indicate that performance remains stable across a wide range of values, suggesting that the framework is not overly sensitive to hyperparameter tuning. This robustness simplifies model training and supports the practical applicability of our approach across datasets with different contrast and noise characteristics.

\myparagraph{Ablation Study on the Loss Components}. Finally, we evaluate the individual and joint contributions of each loss component in~\Cref{loss:sdf_loss}. As summarized in~\Cref{ablation:loss_components}, removing any single term leads to a measurable degradation in performance, whereas combining all components yields the highest Dice accuracy and the lowest boundary error. This confirms that the gradient norm (\(\mathcal{L}_{\nabla}\)), total variation (\(\mathcal{L}_{\text{TV}}\)), and consistency (\(\mathcal{L}_{\text{cons}}\)) terms are complementary: the first two enforce geometric regularity and smoothness, while the latter maintains cross-resolution alignment. Together, they guide the network toward anatomically coherent, high-fidelity segmentations.

\subsection{Scalable Class-Conditional Segmentation (SCCS)}
We further evaluate the effectiveness and scalability of the proposed SCCS strategy through controlled experiments.

\myparagraph{Comparison with Classical Multi-class Segmentation}. \Cref{tab:class_conditioning_comparison} presents a comparison between the proposed SCCS framework and the conventional multi-class segmentation setting, where all class-conditioning channels are processed jointly. The classical all-class model achieves a comparable Dice score (0.771 $\pm$ 0.023 vs. 0.769 $\pm$ 0.016) but requires higher GPU memory (particularly for a high number of classes) and cannot readily generalize to new, previously unseen classes.  In contrast, SCCS focuses on one class at a time, resulting in a constant memory footprint that does not scale with the number of classes. 
This property makes SCCS particularly well-suited for UHR segmentation tasks involving numerous anatomical structures or when hardware resources are limited.

\begin{table}[t!]
\setlength{\tabcolsep}{6pt}
\centering
\scriptsize
\caption{Comparison between all-class conditioning and SCCS.}
\label{tab:class_conditioning_comparison}
\begin{tabular}{lccc}
\toprule
Method & Input Channels & Dice$\uparrow$ & HD95$\downarrow$ \\
\midrule
All-class conditioning & $1 + C$ & 0.771 $\pm$ 0.023 & 0.803 $\pm$ 0.124 \\
SCCS (Ours)            & $1 + 1$ & 0.769 $\pm$ 0.016 & 0.798 $\pm$ 0.117\\
\bottomrule
\end{tabular}
\end{table}

\myparagraph{Generalization to Held-out Classes}. To further assess flexibility and generalization, we evaluate SCCS on unseen anatomical classes. Specifically, one class is excluded during training and later introduced only at test time, as summarized in~\Cref{tab:unseen_generalization}. While the conventional multi-class model performs slightly better on seen classes (0.782 $\pm$ 0.019 vs. 0.771 $\pm$ 0.016), it is fundamentally restricted to the fixed set of labels used during training and cannot infer new structures without retraining. In contrast, SCCS, by design, accepts a class-conditional input that specifies the target class, enabling it to segment previously unseen structures directly. Despite not having encountered the held-out class during training, SCCS achieves a reasonable Dice score of 0.687 $\pm$ 0.036, demonstrating its capacity to generalize across classes. This property is particularly valuable in evolving neuroimaging datasets where anatomical definitions, label sets, or study protocols may expand over time. By decoupling segmentation from fixed label dependencies, SCCS provides a flexible, scalable, and future-proof solution for UHR anatomical segmentation.

\begin{table}[t!]
\setlength{\tabcolsep}{2pt}
\centering
\scriptsize
\caption{Generalization performance on unseen anatomical classes. SCCS enables segmentation of held-out classes without retraining, while the classical model fails to generalize.}
\label{tab:unseen_generalization}
\begin{tabular}{lccc}
\toprule
Method & Input Channels & Seen Classes (Dice $\uparrow$) & Unseen Classes (Dice$\uparrow$) \\
\midrule
Multi-class seg. & $1 + (C-1) $ & \textbf{0.782 $\pm$ 0.019} & N/A \\
SCCS (Ours)                        & $1 + 1$ & 0.771 $\pm$ 0.016 & \textbf{0.687 $\pm$ 0.036}\\
\bottomrule
\end{tabular}
\end{table}

\section{Conclusion}

We introduced a learning-based framework for upscaling 3D segmentations, framing resolution transfer as a general representation learning problem rather than a domain-specific anatomical task. Our method learns to infer HR semantic detail from coarse volumetric labels by predicting continuous signed distance representations, enabling accurate and geometry-aware label refinement without direct supervision at UHRs. Through class-conditional conditioning and scalable architectural design, the approach generalizes across structures and datasets while maintaining computational efficiency. Experiments on HR \emph{ex vivo} MRI demonstrate that the proposed framework bridges the gap between coarse and fine segmentation regimes, offering a scalable path toward high-fidelity 3D label synthesis. More broadly, we see this as a step toward learning-based resolution transfer in structured visual data — extending the idea of label ``super-resolution'' beyond images to the space of semantic 3D geometry.


{
    \small
    \bibliographystyle{ieeenat_fullname}
    \bibliography{main}
}

\clearpage

\maketitlesupplementary

\section{Overview}
\label{sec:overview}
In the supplementary material, we begin with the details of the datasets~\Cref{sec:datasets}, followed by the implementation details in~\Cref{sec:impl_details}. Then, we provide the computational resources in~\Cref{sec:com_res}, followed by a few other qualitative samples in~\Cref{sec:qualitative}. The limitations are provided in~\Cref{sec:limitations}, followed by an analysis on the broader impact in~\Cref{sec:broader_impact} and a statement on the use of LLMs in~\Cref{sec:LLM}.

\section{Dataset Details}
\label{sec:datasets}

\myparagraph{Synthetic data}. As described in the main text, we use synthetic data for training. The UHR isotropic scans are generated from created segmentation labels at a resolution of $\frac{1}{3}$\,mm $\times$ $\frac{1}{3}$\,mm $\times$ $\frac{1}{3}$\,mm. The coarse, LR segmentation $S_{l}$ is obtained using SynthSeg~\cite{billot2023synthseg}, which segments approximately 30 brain regions at 1\,mm isotropic resolution, regardless of the input resolution. We group these regions into 7 foreground classes, \textit{Cortex}, \textit{White Matter}, \textit{Thalamic Mask}, \textit{Pallidum Mask}, \textit{Putamen Mask}, \textit{Caudate and Accumbens}, and \textit{Cerebellar Gray Matter}, along with a background class. The detailed mapping list will be provided to ensure reproducibility.

\myparagraph{Real data}. Following the same class grouping used for the synthetic data, we asked expert annotators to manually label one representative slice from each of 20 real scans in the \textit{U01} dataset~\cite{pesce20223d}.

\section{Implementation Details}
\label{sec:impl_details}
\myparagraph{Network details}. The 3D UNet architecture employed in this paper follows an encoder-decoder structure with skip connections, designed to capture both global context and fine-grained spatial details in volumetric medical images. The encoder consists of four downsampling blocks, each composed of two 3D convolutional layers followed by batch normalization and LeakyReLU activations, with 3D max pooling used to progressively reduce spatial resolution while increasing the number of feature channels. The bottleneck (or bridge) layer connects the encoder and decoder, maintaining the deepest representation with the highest channel dimension. The decoder mirrors the encoder with four upsampling blocks, where each block begins with a transposed 3D convolution to upsample the feature map, followed by concatenation with the corresponding encoder features (skip connection), and two additional convolutional layers with normalization and activation. The final output layer is a 3D convolution that maps the feature maps to the desired number of output channels. This design enables precise voxel-wise predictions while maintaining spatial consistency across the 3D volume. The codes will be released upon acceptance to ensure reproducibility.

\section{Computational Resources}
\label{sec:com_res}

The experiments are conducted on an NVIDIA A40 GPU (48GB), using a 26-core Intel(R) Xeon(R) Gold 6230R CPU @ 2.10GHz and 200 GB RAM.

\section{Qualitative Results}
\label{sec:qualitative}

\begin{figure*}[t]
    \centering
     \setlength{\tabcolsep}{0.5pt}
    \begin{tabular}{ccccccc}
    \includegraphics[width=.125\textwidth]{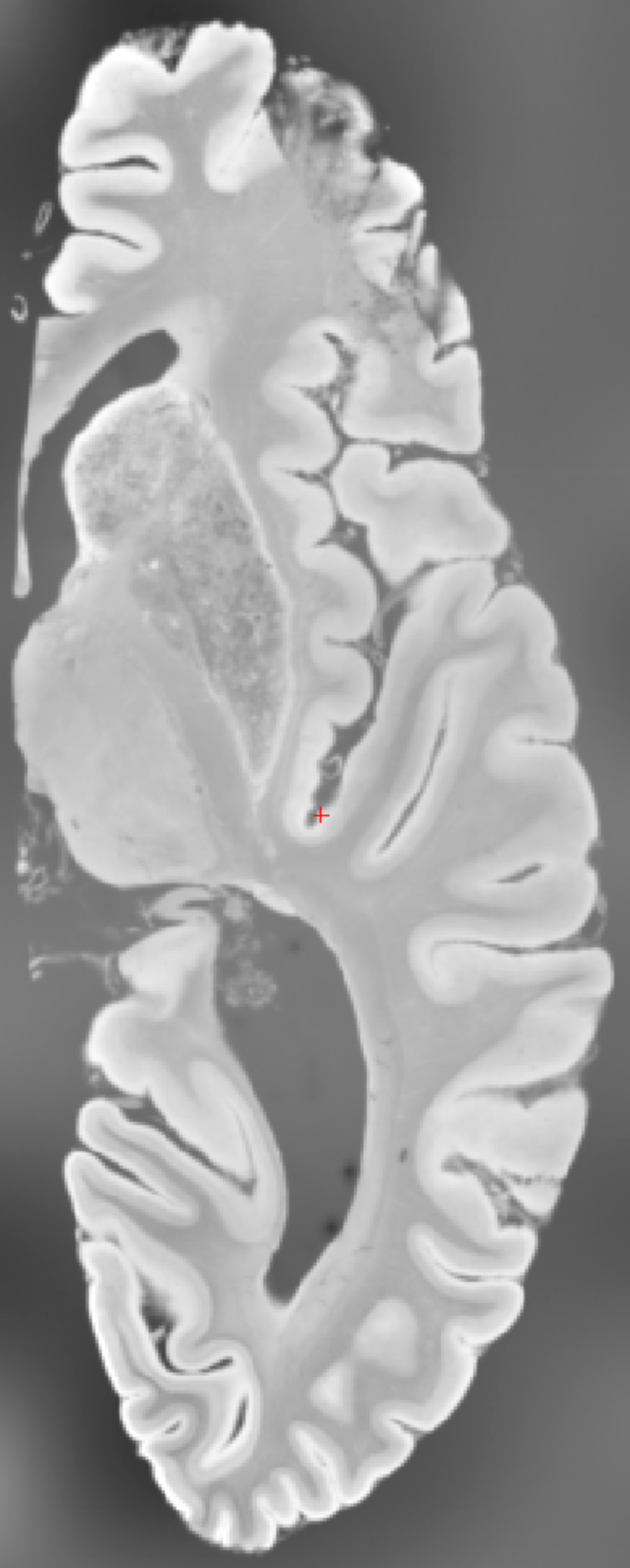} &
    \includegraphics[width=.125\textwidth]{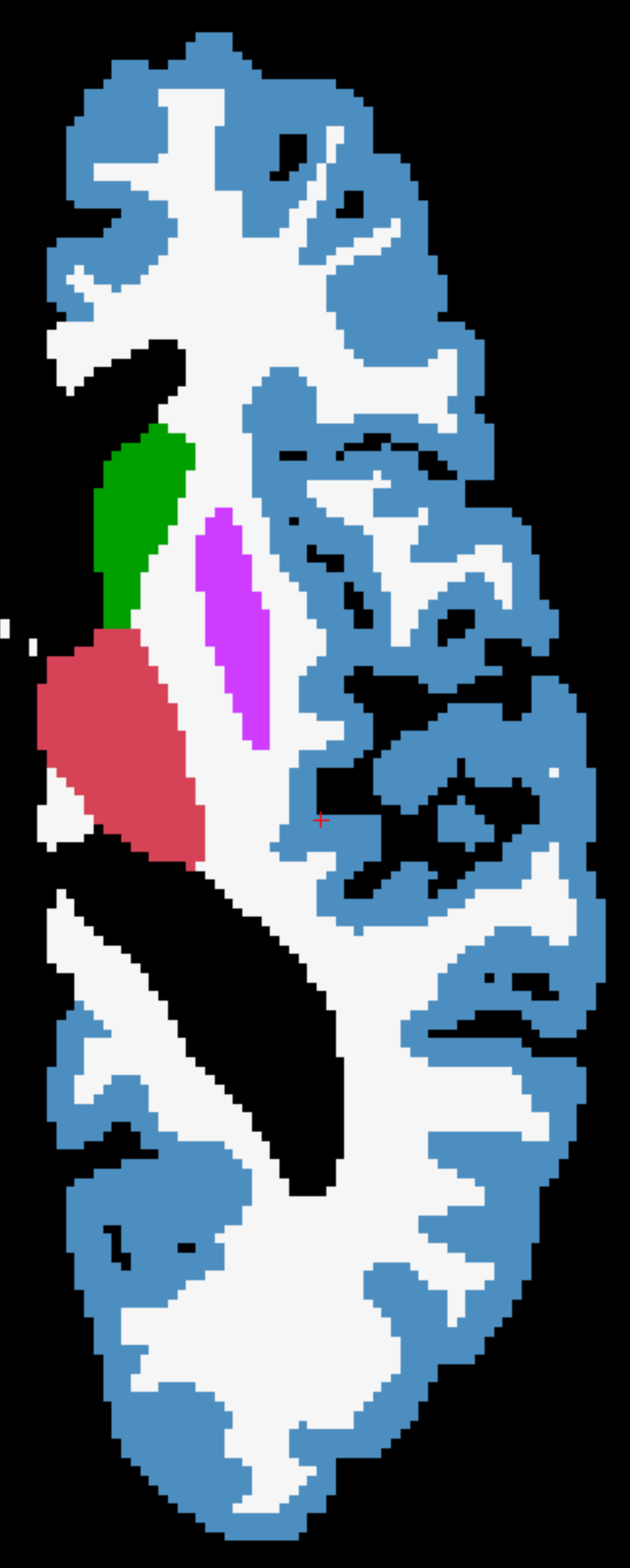} &
    \includegraphics[width=.125\textwidth]{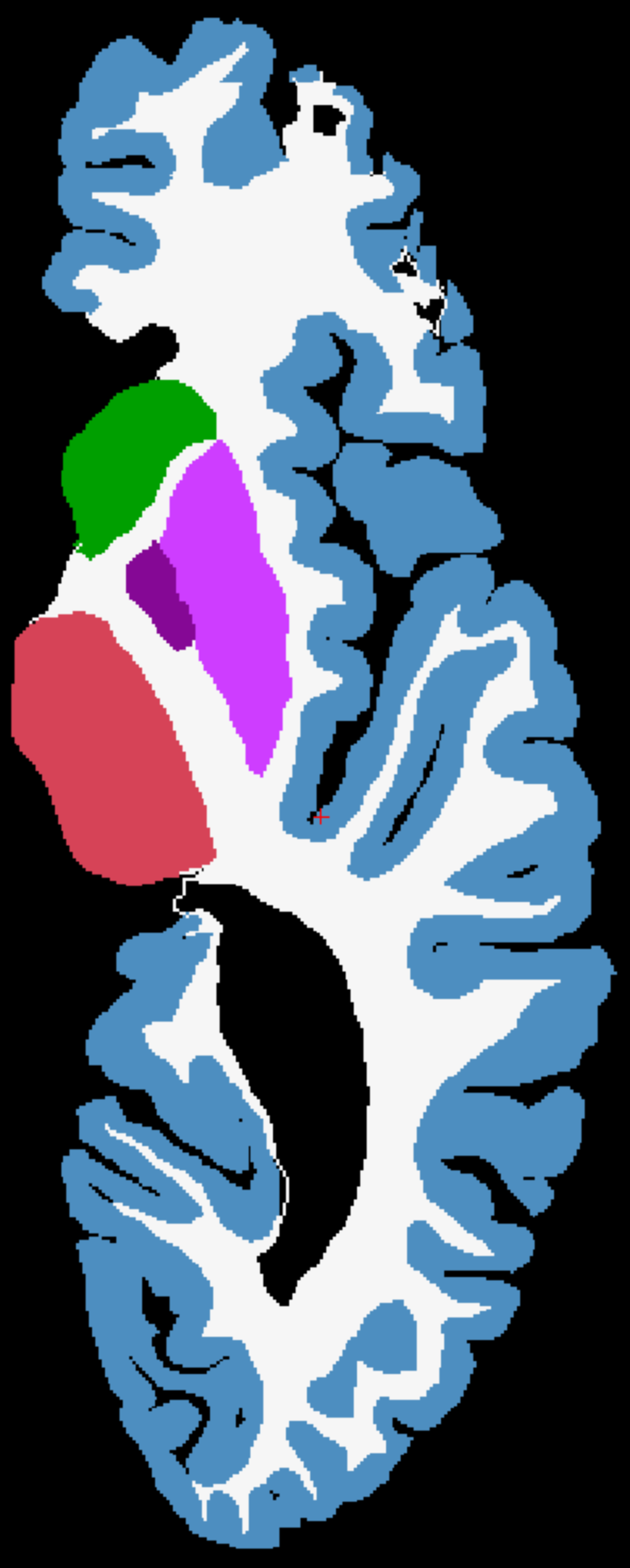} &
    \includegraphics[width=.125\textwidth]{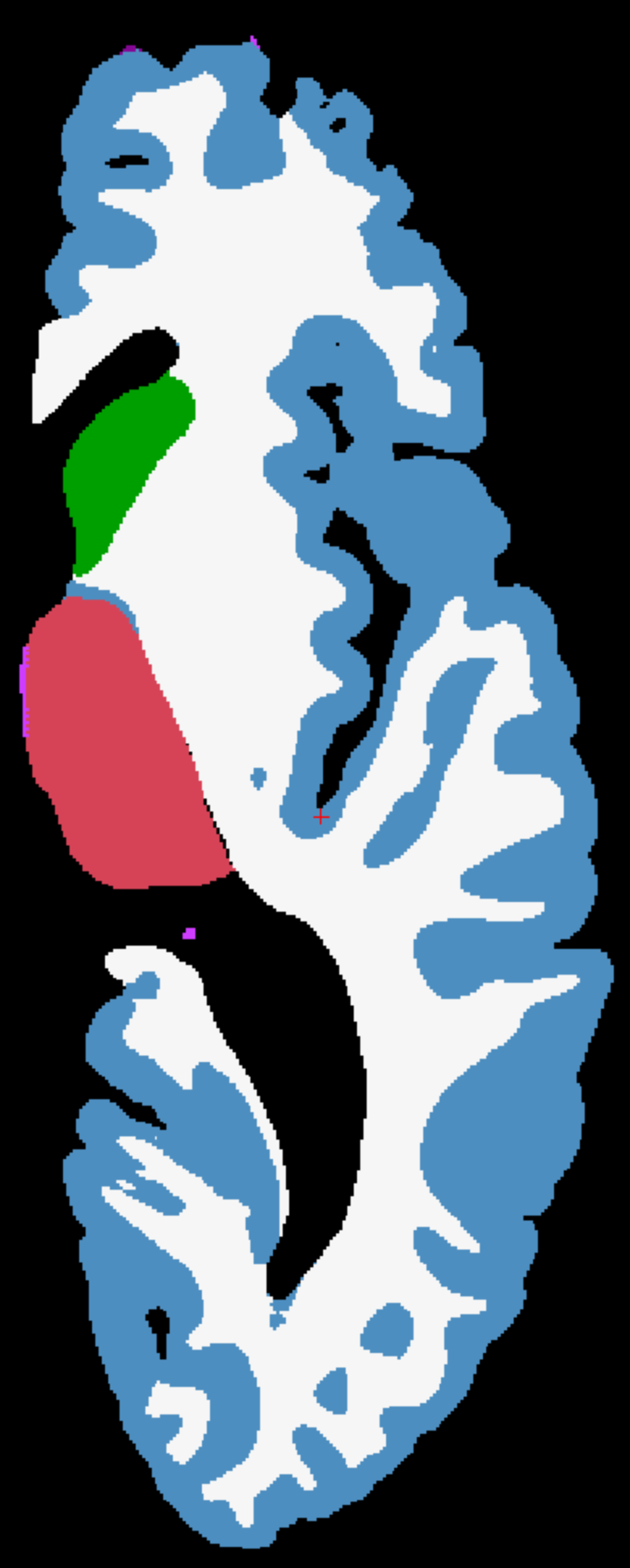} &
    \includegraphics[width=.125\textwidth]{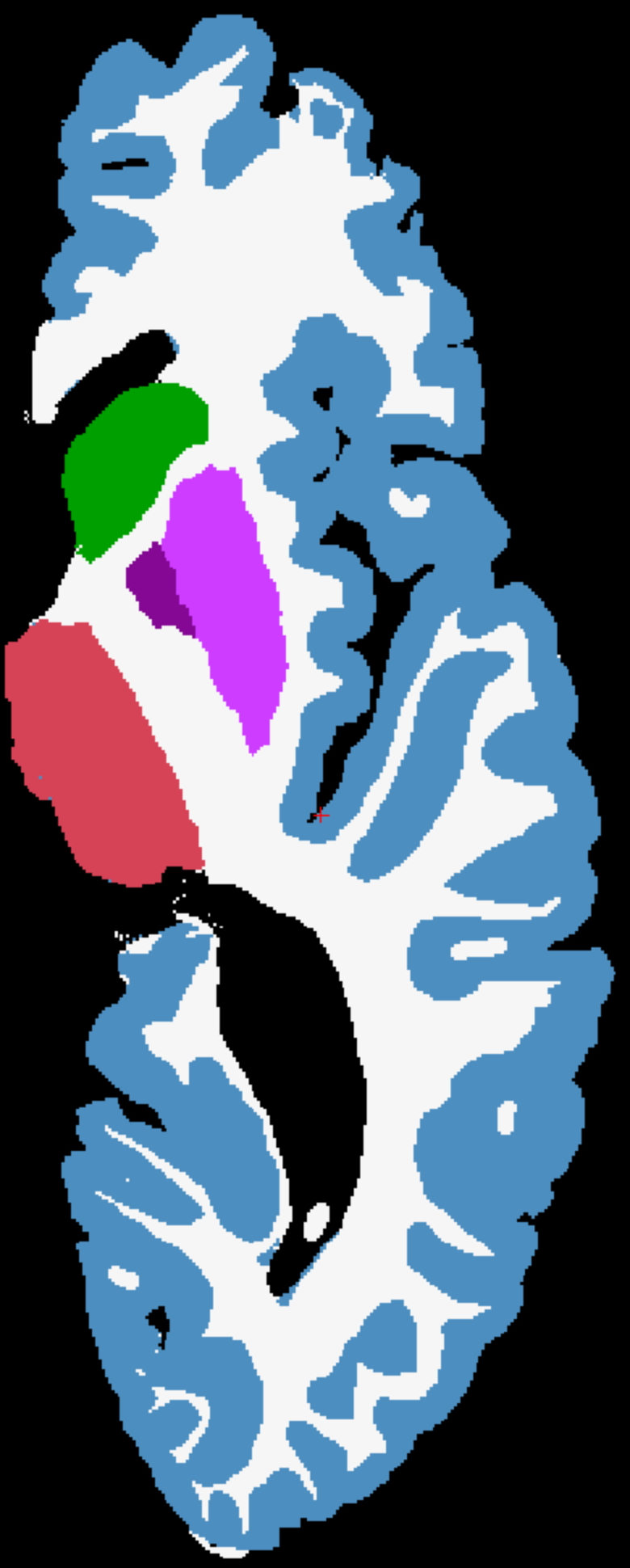} &   
    \includegraphics[width=.125\textwidth]{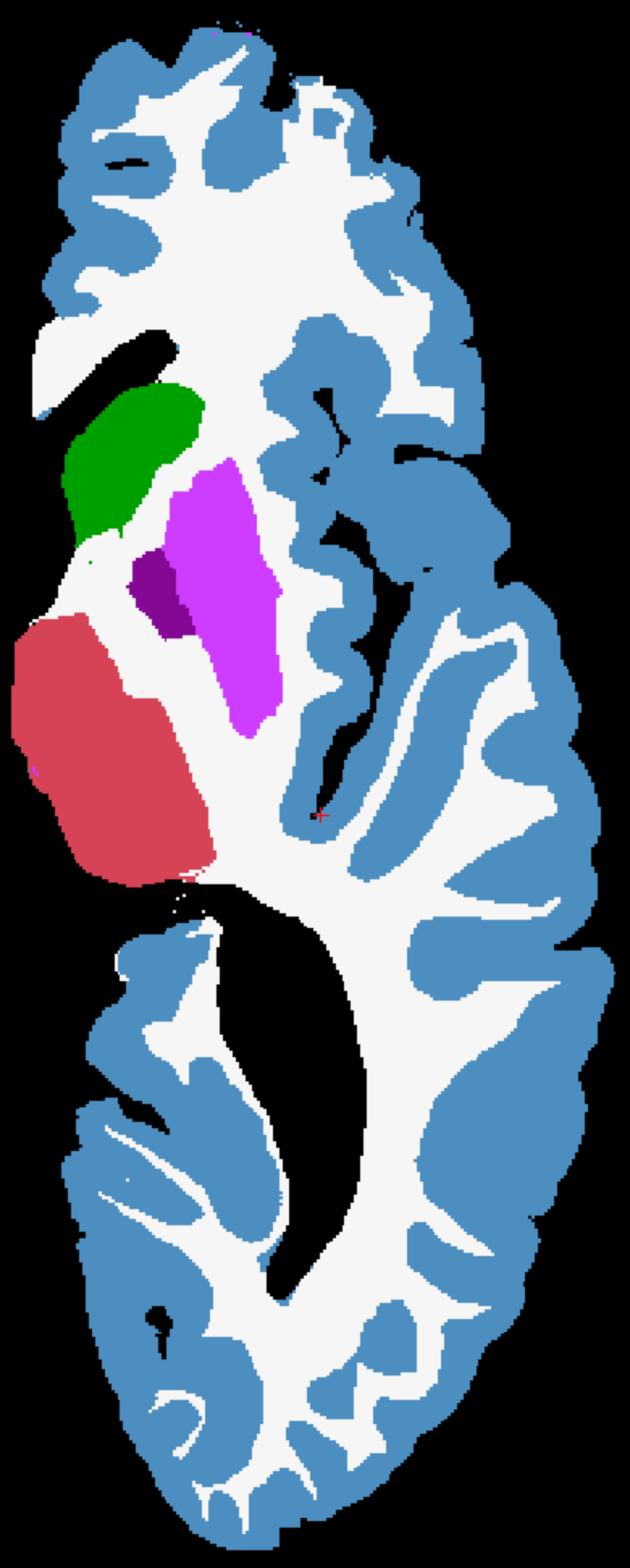} &
    \includegraphics[width=.125\textwidth]{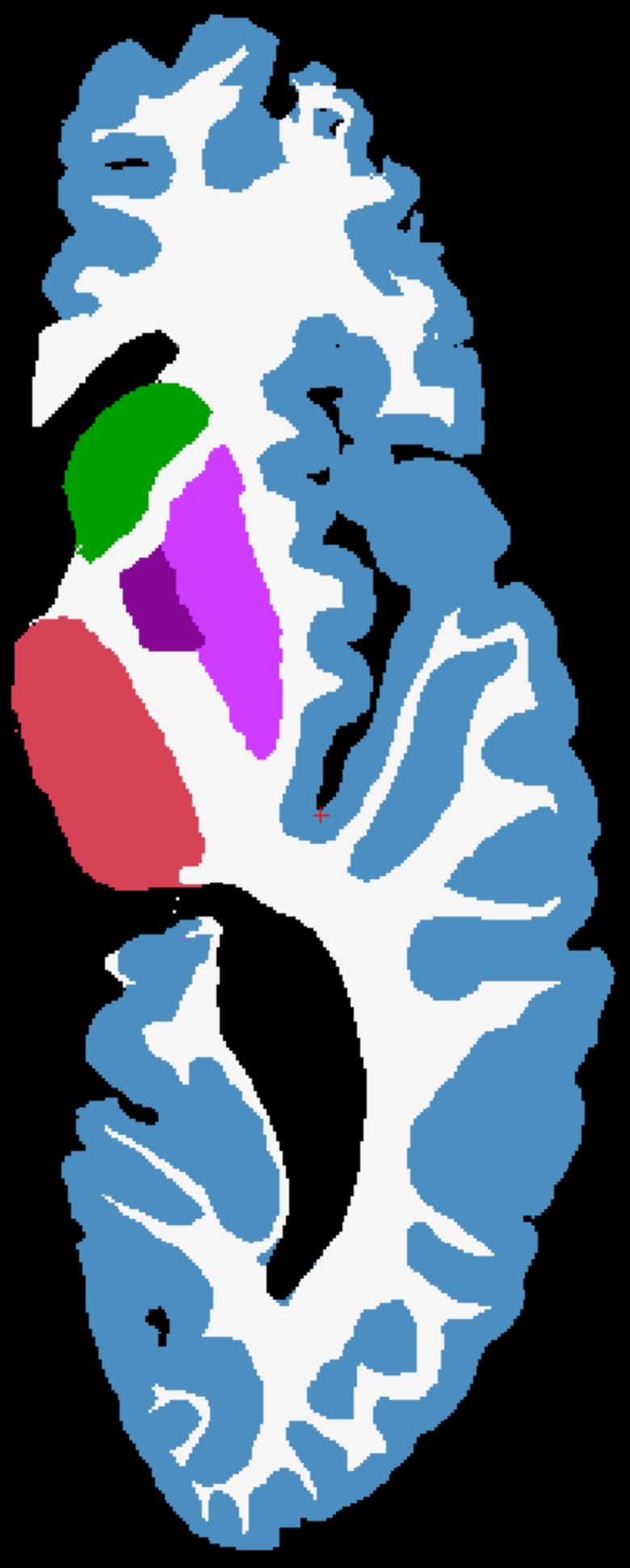} \\
     (a) Input & (b) LR guidance & (c) GT & (d) Naive Seg & (e) Predict $\hat{S}_H$ & (f) Regress $\hat{\phi}_H$ & (g) SCCS  \\
    \end{tabular}
    \caption{Qualitative results. (a-c) show the input, LR guidance, and ground truth (GT). (d-g) show segmentations with different methods. }
    \label{fig:qualitative_real}
\end{figure*}

For qualitative results, we provide another sample from \textit{U01} in~\Cref{fig:qualitative_real}.

\section{Limitations}
\label{sec:limitations}

A key limitation of the proposed method lies in its reliance on automatic LR segmentations as spatial guidance. While this strategy significantly reduces manual labeling effort, it inherently assumes that these coarse labels are accurate and spatially consistent. In practice, however, these LR segmentations may contain systematic biases or anatomical imprecision. These imperfections can propagate through the network, potentially leading to degraded segmentation performance at higher resolutions. One possible solution is to incorporate uncertainty modeling or confidence-weighted supervision, where the model learns to discount or correct for less reliable regions in the coarse labels. Additionally, leveraging self-supervised refinement mechanisms that iteratively improve the alignment between LR and HR outputs could further mitigate this issue.

Another important limitation is the limited validation on real, fully annotated 3D clinical datasets. Although the paper demonstrates strong performance on synthetic data and sparsely labeled real data (e.g., single-slice annotations), the generalizability of the proposed approach to densely annotated, HR clinical scans remains uncertain. This is particularly relevant given the variability in acquisition protocols, scanner hardware, and anatomical differences across patient populations. To strengthen the empirical evidence and assess robustness, future work should include comprehensive benchmarking on public and private datasets with full volumetric annotations (e.g., HCP, OASIS, ADNI). Incorporating domain adaptation techniques or semi-supervised learning frameworks could also help bridge the gap between synthetic and clinical domains, further enhancing the method's practical utility in real-world settings.

\section{Broader Impact}
\label{sec:broader_impact}

The broader impact of our work lies in its potential to democratize access to detailed, high-fidelity brain image analysis without the prohibitive cost of dense manual annotations. By leveraging LR coarse labels and a scalable, class-conditional framework, the proposed method makes it feasible to segment UHR brain MR scans, which are increasingly used in neuroscience and clinical research, using limited supervision and computational resources. This can accelerate research in neurodegenerative diseases, brain development, and population-level studies where large-scale, accurate segmentation is essential. 

Additionally, the framework’s ability to generalize to unseen classes and operate efficiently in memory-constrained settings makes it adaptable to LR clinical environments or global health applications. However, as with any medical AI tool, careful validation is essential to avoid biases or errors introduced by synthetic or weak labels. If responsibly developed and adopted, the method could contribute meaningfully to advancing scalable, accessible, and precise neuroimaging analysis.

\section{Usage of LLM}
\label{sec:LLM}

We only use LLM to improve the writing quality and grammar check of the manuscript.

\end{document}